\documentclass[default,iicol,Numbered]{sn-jnl}
\onecolumn
\usepackage{amssymb}
\usepackage{amsmath}
\usepackage{amsmath,amsfonts}
\usepackage{algorithmic}
\usepackage{algorithm}
\usepackage{array}
\usepackage{threeparttable}   
\makeatletter
\newenvironment{breakablealgorithm}
  {
   \begin{center}
     \refstepcounter{algorithm}
     \hrule height.8pt depth0pt \kern2pt
     \renewcommand{\caption}[2][\relax]{
       {\raggedright\textbf{\fname@algorithm~\thealgorithm} ##2\par}%
       \ifx\relax##1\relax 
         \addcontentsline{loa}{algorithm}{\protect\numberline{\thealgorithm}##2}%
       \else 
         \addcontentsline{loa}{algorithm}{\protect\numberline{\thealgorithm}##1}%
       \fi
       \kern2pt\hrule\kern2pt
     }
  }{
     \kern2pt\hrule\relax
   \end{center}
  }
\makeatother
\usepackage[caption=false,font=normalsize,labelfont=sf,textfont=sf]{subfig}
	
	\usepackage{textcomp}
	\usepackage{stfloats}
	\usepackage{url}
	\usepackage{verbatim}
	\usepackage{graphicx}
	\usepackage{multirow}  

	\usepackage[T1]{fontenc} 
	\usepackage{xcolor}
	\definecolor{myblue}{RGB}{70,130,180}
	\definecolor{myorange}{RGB}{255,165,0}
	\usepackage{tikz} 
	\usetikzlibrary{positioning}
	\DeclareUnicodeCharacter{200B}{}
\usepackage{tabularx} 
\usepackage{ragged2e} 
\usepackage{booktabs} 
\usepackage{rotating} 
\usepackage{threeparttable}
\usepackage{makecell} 
\usepackage[numbers]{natbib}

\theoremstyle{thmstyleone}%
\theoremstyle{thmstyletwo}%

\theoremstyle{thmstylethree}%

\begin{document}

\title[Article Title]{UFGraphFR: Graph Federation Recommendation System based on User Text description features}


\author*[1]{\fnm{Wang} \sur{Xudong}}\email{wangxudong@stud.tjut.edu.cn}

\author[1]{\fnm{Hao} \sur{Qingbo}}\email{xuyan@stud.tjut.edu.cn}


\author[1]{\fnm{Xiao} \sur{Yingyuan}}\email{yyxiao@tjut.edu.cn}

\affil*[1]{\orgdiv{School of Computer Science and Engineering}, \orgname{Tianjin University of Technology}, \orgaddress{\street{391 Binshui West Road}, \city{Xiqing}, \postcode{300384}, \state{Tianjin}, \country{Tianjin}}}



\abstract{Federated learning provides a privacy-preserving framework for recommendation systems by enabling personalized recommendations. However, the growing scale of user data and the demand for real-time, high-quality recommendations have created unprecedented computational burdens, necessitating support from supercomputing infrastructure. Traditional federated recommendation methods treat each user as an isolated entity, failing to construct a global user relationship graph that captures collaborative signals, thereby limiting recommendation accuracy. To address this limitation, this paper proposes UFGraphFR, a graph-based federated recommendation system leveraging user textual description features. Its core innovation lies in proposing a novel paradigm that uses users’  textual attributes as proxy signals to securely construct a global user relationship graph on the server side without requiring raw interaction data alignment or upload, thereby providing an architectural foundation for privacy preservation. Specifically, the framework comprises three key technical components: 1) On the client side, private structured data is first transformed into textual descriptions via prompt templates and encoded into semantic vectors using a pre-trained model; a trainable linear layer, driven by local interaction data, further maps static semantic vectors into dynamic, personalized low-dimensional user embeddings. 2) On the server side, based solely on model weights uploaded by clients (rather than raw data), a user relationship graph is securely reconstructed by computing the semantic similarity of user embedding weights, and a lightweight graph neural network is employed for information propagation to aggregate global knowledge. 3) On the client side, user behavior sequences are modeled individually using a Transformer architecture to capture long-term interest dependencies. This design not only effectively addresses the issue of missing collaborative signals caused by user data isolation in federated recommendations but also offloads computationally intensive graph construction and aggregation tasks to the central server, aligning with a hybrid computing model that leverages high-performance computing (HPC) clusters to handle massive, real-time privacy-preserving recommendation tasks. Extensive experiments on four benchmark datasets demonstrate that UFGraphFR significantly outperforms state-of-the-art federated and centralized baseline methods in terms of recommendation accuracy (HR@10) and personalization degree (NDCG@10). The framework remains robust across different pre-trained models and shows potential for balancing privacy and performance when optional noise is introduced as a further mitigation. This work provides a practical path for deploying computationally intensive, privacy-preserving recommendation tasks in supercomputing environments, bridging the gap between advanced federated learning and the scalable, high-throughput requirements of supercomputing environments. The code is available at: \texttt{https://github.com/trueWangSyutung/UFGraphFR}.}




\maketitle

\section{Introduction} 
In the era of big data and privacy computing, intelligent recommendation systems have become an indispensable infrastructure for modern digital services, enabling personalized movie suggestions, targeted food delivery recommendations, and even emotion-aware content curation through user comment analysis\cite{zhang_gpfedrec_2024}. However, conventional centralized paradigms that aggregate raw user data on servers pose critical privacy risks, as evidenced by growing user reluctance to share personal information with service providers\cite{yuan_hetefedrec_2023}. Federated learning emerges as a promising privacy-preserving framework, allowing collaborative model training without exposing users' private data \cite{mcmahan_communication-efficient_2023}. Its application in recommender systems enables the construction of complete models while keeping raw data localized on client devices\cite{chai_secure_2021,10.24963/ijcai.2023/851}. 

However, a fundamental limitation of existing federated recommendation approaches stems from the isolation of raw user-item interaction records on local devices, which prevents the server from accessing the global user-item interaction matrix\cite{he_lightgcn_2020, wang_learning_2021}. Consequently, unlike traditional graph neural network recommendation models that leverage global graph structures (such as user-item bipartite graphs or user social graphs) to capture complex user correlations and collaborative signals, federated systems primarily rely on limited local data from individual users. This constraint inherently caps the upper bound of recommendation accuracy, as models cannot effectively exploit the rich relational patterns available in centralized settings.

In this paper, we propose a novel framework for personalized federated recommendation, termed \textbf{UFGraphFR} (\textbf{U}ser text-description \textbf{F}eature \textbf{Graph} \textbf{F}ederated   \textbf{R}ec), which addresses a key challenge: how to find a proxy signal that enables the server to measure user similarity and approximately build a 'user relationship graph' without exposing raw private data. To solve this problem, UFGraphFR securely transforms local private structured attributes into textual descriptions and maps them into a unified semantic space using pre-trained language models, thereby enabling safe user relationship construction. The main contributions of this work are summarized as follows:

\begin{itemize}
 \item We innovatively leverage user text descriptions as a secure bridge for constructing user relation graphs, breaking away from ID-centric paradigms and enabling semantic-aware similarity modeling under privacy constraints.
 \item We design a learning framework UFGraphFR: on the client side, private structured data is first converted into text descriptions and encoded into semantic vectors using pretrained models; on the server side, user relation graphs are reconstructed securely using aggregated model weights without accessing raw data, followed by information propagation via lightweight graph neural networks.
 \item Through comprehensive experiments on four benchmark datasets, we validate the superiority of UFGraphFR in recommendation accuracy, personalization capability, and robustness, supported by in-depth ablation studies.
\end{itemize}

The rest of this paper is organized as follows: Section II reviews the related work on federated recommender systems and graph-based learning recommender systems. Section III introduces the UFGraphFR method proposed in this paper, including its overall framework and key technical details. Section IV will focus on the experimental part, including performance experiments and ablation experiments.

\section{Related Work}

\subsection{Federated Recommendation Systems}

\begin{figure}[ht!]
	\centering
	\includegraphics[width=0.8\linewidth]{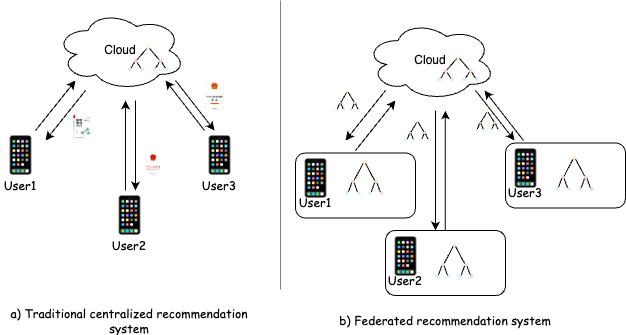}
	\caption{a) shows the traditional centralized recommendation system. The user client first upload its own user characteristics and interactive data (with or without the consent of the user) to the server, and the server trains a recommendation model in the cloud to recommend suitable items for the user. b) shows a federated learning framework. Each client has a local model. The cloud sends the model parameters to the client in advance, and the client conducts local training without uploading private data to the cloud and uploals the trained model parameters to the server for aggregation.}
	\label{fedrec}
\end{figure}

Federated Recommendation Systems (FedRS, as shown in Fig.\ref{fedrec}) is an important extension of Federated Learning (FL) in the field of recommendation, which realizes cross-client personalization modeling by protecting the user's privacy. Personalized modeling, allows the user's raw data to complete the training process of the Recommender system(RecSys) without leaving the client while protecting the user's privacy.

Some existing basic recommender system approaches (e.g., Matrix-Factorization-Based RecSys\cite{2009Matrix}, Neural Collaborative Filtering Based RecSys\cite{10.1145/3038912.3052569}) have been combined with the FL framework to form, for example, matrix factorization based federated recommender systems  FCF\cite{ammaduddin2019federatedcollaborativefilteringprivacypreserving}, FedMF\cite{chai_secure_2021}, MetaMF\cite{lin_meta_2020}, FedRecon\cite{singhal_federated_2022}, and neural collaborative filtering based federated recommender system FedNCF\cite{10.1016/j.knosys.2022.108441}. After this, more and more FedRSs were proposed. Zhang et al.\cite{zhang2023dual} proposed a personalized FedRec framework, PFedRec, which removes user embedding and learns personalized scoring functions to capture user preferences. However, this framework ignores the correlation between users and does not respond well to the common preferences among similar users. To compensate for this shortcoming of correlation between users, Liu et al.\cite{liu2022federated} enhanced the federated recommender system by introducing the user's social network. However, this framework exposes the user's social network explicitly, which goes against the premise of protecting user privacy. Wu et al.\cite{wu_federated_2022} proposed FedPerGNN, an approach to capture user-item correlations using local map neural networks. At the same time, FedPerGNN also uses servers to find higher-order neighbors, thus providing more favorable information for local model training. Chen et al.\cite{10988883} propose FedSI, a Bayesian deep neural network-based subnetwork inference framework for personalized federated learning. By selecting parameters with high variance for Bayesian inference and fixing the rest deterministically, FedSI efficiently quantifies systematic uncertainty while outperforming existing baselines in heterogeneous federated learning scenarios. Chen et al.\cite{CHEN2026104454} propose FedPP, a federated neural nonparametric point process model that integrates neural embeddings with Sigmoidal Gaussian Cox Processes (SGCPs) on the client side and employs a divergence-based distribution aggregation mechanism. This approach effectively models sparse events and uncertainties while preserving privacy in federated settings. However, this approach aligns historical interactions with the server leads to high computational overhead, and increases the risk of exposing users' private information.

Therefore, correlation among users is a major pain point in federated recommender systems at this stage, and thus many researchers began to target FedRS frameworks with graph-guided mechanisms. Zhang et al.\cite{zhang_gpfedrec_2024} proposed GPFedRec, a FedRS framework based on a graph-guided mechanism, GPFedRec designs a graph-guided aggregation mechanism to capture user preference correlations. Wang et al.\cite{wang2024clusterenhancedfederatedgraphneural} proposed CFedGR, a clustering-enhanced federated graph neural network recommender system, CFedGR proposes a clustering technique to enhance the user's local first-order interaction subgraph. Wang et al.\cite{wang2025p4gcnverticalfederatedsocial} proposed P4GCN, a Two-Party Graph Convolutional Network implementation of a vertical federated recommender system, which utilizes privacy-preserving Two-Party Graph Convolutional Networks (P4GCN) to improve the accuracy of recommendations without direct access to sensitive social information. So, the main direction of graph-guided FedRS at this stage is to construct user-user association graphs or user-item graphs without violating privacy and apply them to the whole federation training process.

However, FedRSs at this stage are focused on federated recommender systems based on the ID paradigm, which are limited in their semantic expressiveness, resulting in insufficient generalization ability and cold-start adaptation, as sufficient training data may not be available in such scenarios. Therefore, inspired by the success of Language Models (LMs) and their strong generalization ability, the potential of LMs can be exploited to enhance the capabilities of recommender systems. Luo et al.\cite{ren2024easyrecsimpleeffectivelanguage}. proposed EasyRec, a centralized recommender system that employs a text-behavior alignment framework that combines contrastive learning with collaborative language model tuning to ensure tight alignment between the text-enhanced semantic space and collaborative behavioral information. Zhang et al.\cite{zhang2024federatedrecommendationmeetscoldstart} proposed FedRS for IFedRec, which learns two sets of item representations by simultaneously utilizing item attributes and interaction records and designed an item representation alignment mechanism for aligning the two item representations within the FL framework and learning a network of meta-attributes on the server.

Inspired by the above, we combine graph-guided FedRS and LM-based FedRS to design a novel approach to build a user relationship graph, which utilizes users' meta-attributes and pre-trained language models (PLMs) to construct a relationship graph between users. The aggregation of model parameters is achieved through this graph.

Another significant challenge in federated learning is client selection—specifically, determining how many clients participate and how they are chosen. Common approaches include random selection and full client participation. Random selection is straightforward, widely adopted in existing FedRSs, and highly efficient when scaling to a large number of clients. \cite{liu2023privaterec,10.1145/3548456,9626622,wu2022federated,10.1145/3511808.3557320,10.1145/3501815}
However, this method ignores the heterogeneity of clients, leading to suboptimal model convergence, as some clients may provide noisy or biased updates, which reduces the convergence of the model. Whereas Select All \cite{ammad2019federated,10.1145/3442381.3449926,minto2021stronger,chai2020secure,lin2020fedrec,liang2021fedrec++}  is an approach where all clients participate in each round of training, this approach has less efficient communication and is not suitable for devices with a large number of clients, but is often used in cross-platform FedRS approaches. So in our model, we use the full selection method.

\subsection{Graph Learning-based Recommendation System}

Recommendation systems for graph learning, as an emerging paradigm, are enhanced user (item) embeddings that learn by explicitly utilizing neighbor information in the graph structure. A common strategy is to integrate user-item interaction diagrams into a collaborative filtering framework. He et al. proposed a model called LightGCN\cite{he_lightgcn_2020}, which applies graph convolutional networks to user-item interaction graphs to enrich representation learning for user preference prediction. We can also consider the adjacencies between all the items and form an item interaction graph.  In sequence recommendation, it is also a good way to enhance learning through sequence diagrams\cite{latifi_streaming_2022}. With the rise of social networks, social recommendation based on social networks (the relationship network formed by using the social relationship between users) to enhance modeling gradually emerged\cite{Wu_2019}. 
\begin{table*}[ht]
\small
    \caption{Comparison of related work.}

    \label{tab:comparison}
    \begin{tabular*}{1.0\linewidth}{p{1.8cm} p{4cm} p{4cm} p{4.5cm}}
\toprule
\textbf{Method} & \textbf{Core Idea} & \textbf{Pros} & \textbf{Cons/Limitations} \\
\midrule
\textbf{FedMF \cite{chai_secure_2021}} & Applies matrix factorization within the FL framework. Users train locally and upload gradients. & Simple and efficient; Establishes a baseline for FL in recommendation. & Ignores user-user correlations; Limited by the linear nature of MF. \\
\addlinespace

\textbf{FedNCF \cite{10.1016/j.knosys.2022.108441}} & Extends Neural Collaborative Filtering to FL, using MLPs for non-linear user-item interaction modeling. & Captures non-linear interactions; More expressive than linear models like FedMF. & Still treats users in isolation; High communication cost due to large MLP parameters. \\
\addlinespace

\textbf{PFedRec \cite{zhang2023dual}} & Removes user embeddings from the server, learning personalized scoring functions on each client. & Achieves high personalization; Decouples user and item learning. & Fails to model collaborative signals between similar users; Performance may degrade for users with sparse data. \\
\addlinespace

\textbf{FedRecon \cite{singhal_federated_2022}} & Retrains user embeddings in every communication round based on local data. & Enhances personalization by avoiding stale embeddings. & Computationally expensive for clients; May lead to unstable convergence. \\
\addlinespace

\textbf{FedPerGNN \cite{wu_federated_2022}} & Deploys GNNs on clients and uses a graph extension protocol to find higher-order neighbors. & Leverages structural information for better recommendations. & Requires aligning historical interactions with the server, risking privacy leaks; High computational and communication overhead. \\
\addlinespace

\textbf{GPFedRec \cite{zhang_gpfedrec_2024}} & Proposes a graph-guided aggregation mechanism to capture user preference correlations. & Explicitly models user-user relationships; Improves performance by leveraging neighborhood information. & The method for constructing the user graph is not explicitly detailed in the provided text, potentially relying on sensitive data. \\
\addlinespace
\textbf{EasyRec \cite{ren2024easyrecsimpleeffectivelanguage}} & A centralized system using a text-behavior alignment framework with contrastive learning and language model tuning. & Effectively integrates rich semantic information from text; Strong generalization capability. & Centralized architecture violates user privacy; Not applicable to privacy-sensitive federated settings. \\
\bottomrule
\end{tabular*}
\end{table*}

In short, there is substantial work remains to be done to learn user embedding from user project interaction diagrams or social networks, or even to aggregate the two into a unified graph to enhance user representation. However, at a time when existing methods are increasingly focused on user privacy, this centralized recommendation system, which requires users to upload data to the cloud for training, is high-risk (as shown in Figure \ref{fedrec}.a) because its centralized access to user data violates user privacy\cite{zhang_gpfedrec_2024}. As a result, federal recommendation are gradually emerging, which combine privacy protection technologies to protect users' private data from being leaked. 

To provide a systematic overview of the existing federated recommendation landscape, Table~\ref{tab:comparison} summarizes and compares the key characteristics, advantages, and limitations of several representative methods discussed in this section.

\section{Preliminary}

\subsection{Federated Recommendation}  
The $U$ is  user sets and the $I$ is  item sets, respectively, let $r_{ui}$ be user-item interaction data between user $u$ and item $i$. Here is a recommendation system model $f$ for parameter $\theta$, which predicts $\hat{y_{ui}}=f(u,i|\theta)$ for users $u$ and $i$.  On the central server, we represent the graph between all users with $\mathcal{G(U,E})$, where $\mathcal{U}$ represents the set of users and  $\mathcal{E}$ represents the set of edges. At the same time,  $\mathcal{A}={\{0,1\}}^{N\times N}$ represents its corresponding adjacency matrix form, and N represents the total number of users in the user set $\mathcal{A}_{ui} = 1$ indicates that $u$ and user i are associated.

For the above model, the purpose of the federated recommendation system is to predict $u$'s preference for item i as $\hat{y_{ui}}=f(u,i|\theta^*)$ , and the optimal model parameter at this time is $\theta^*$, as shown in formula \eqref{con:fr}. 
\begin{equation}
	\theta^* = argmin_{\theta} = \sum_{i=1}^{N} \omega_i \mathcal{L}_i(\theta) \label{con:fr}
\end{equation}
where $\mathcal{L}_i(\theta)$ is the loss of the local client participating in the training, and the parameter $\theta^*$ is learned by minimizing the local loss of all clients with the client weight $ \omega_i$.

For clarity, we summarize the key notations used throughout this paper in Table~\ref{tab:notation}. These symbols are fundamental to the mathematical formulation and description of the proposed UFGraphFR framework.

\begin{table}[!ht]
\caption{Summary of key notations.}
\label{tab:notation}
\centering
\begin{tabular*}{\linewidth}{c p{12cm}}
\toprule
\textbf{Symbol} & \textbf{Definition} \\
\midrule
$U, N$ & Set and number of users. \\
$I$ & Set of items. \\
$r_{ui}$ & Interaction between $u$ and  $i$. \\
$f(\theta)$ & Recommendation model with parameters $\theta$. \\
$\theta_{\text{global}}, \theta_{\text{local}}$ & Global shared and local private model parameters. \\

$\theta_{\text{global}}, \theta_{\text{local}}$ & Global shared and local private model parameters. \\
$\theta_{\text{umlp}}, \theta_{\text{transformer}}, \theta_{\text{score}}$ & User Feature Refinement MLP, Temporal Transformer module  and Predictive Scoring Function parameters. \\

$\mathcal{G(U,E)}$ & User relationship graph with adjacency matrix $\mathcal{A}$. \\
$\mathcal{L}_i(\theta)$ & Local loss on client $i$. \\
$\omega_i$ & Client $i$'s aggregation weight. \\
$\mathcal{P}_u), v_u$ &  $u$'s attribute and PLM-generated text embedding. \\
$e_u$ & Final low-dimensional user embedding. \\
$d, d_1, d_2$ & Dimensions: joint embedding, PLM output, linear layer output. \\
$L$ & Length of user's interaction sequence. \\
$e_{i_t}$ & Embedding of the $t$-th interacted item. \\
$d_{\text{item}}$ & Item embedding dimension. \\
$K_h, d_h$ & Number and size of Transformer attention heads. \\
$\eta, \lambda$ & Learning rate and regularization coefficient. \\
$\epsilon$ & Noise scale for differential privacy. \\
\bottomrule
\end{tabular*}
\end{table}

\subsection{Key Assumptions}
Based on the insight that users with similar semantic profiles tend to exhibit similar behavioral preferences, we introduce the following core assumption: the similarity between users' text descriptions, which are derived from their structured attributes and encoded via pre-trained language models (PLMs), can serve as a reliable proxy for their preference similarity. Formally, for two users $u$ and $v$, we posit that the cosine similarity $S_{uv}$ between their respective text-derived semantic vectors correlates positively with the similarity of their interaction patterns and item preferences.

To operationalize this assumption within the federated learning framework while preserving privacy, we leverage the user-specific weight matrix $W_u$ from the joint embedding layer as the secure representation for constructing the user relationship graph. Specifically, $W_u \in \mathbb{R}^{d_1 \times d_2}$ is the trainable linear projection layer parameter that adapts the static PLM-generated embedding $v_u$ to the user's local interaction data $D_u$. This matrix dynamically encodes both the user's inherent semantic attributes and their evolving behavioral preferences. Thus, it is hypothesized that the similarity $S_{uv}$ calculated between the flattened vectors of $W_u$ and $W_v$ reflects the underlying preference similarity between users $u$ and $v$, allowing the server to approximate a global user relationship graph without accessing raw private data. 

\section{Methodology}
In this section, we present the Graph Federation Recommendation System based on User Text description features (UFGraphFR). There are four steps in each communication round: 
\begin{itemize}
	\item \textbf{Local Training}: Clients initialize project embeddings using global parameters and train recommendation models with private interaction data.
	\item \textbf{Parameter Uploading}: Clients transmit user joint embedding weights and local item embeddings to the server.
	\item \textbf{Graph Aggregation}: The server constructs user relation graphs from text embeddings and aggregates parameters through graph convolution.
	\item \textbf{Global Distribution}: Updated global item embeddings are broadcast to all clients for next-round initialization.
\end{itemize}

\begin{figure*}[hbpt]
	\centering
	\includegraphics[width=\linewidth]{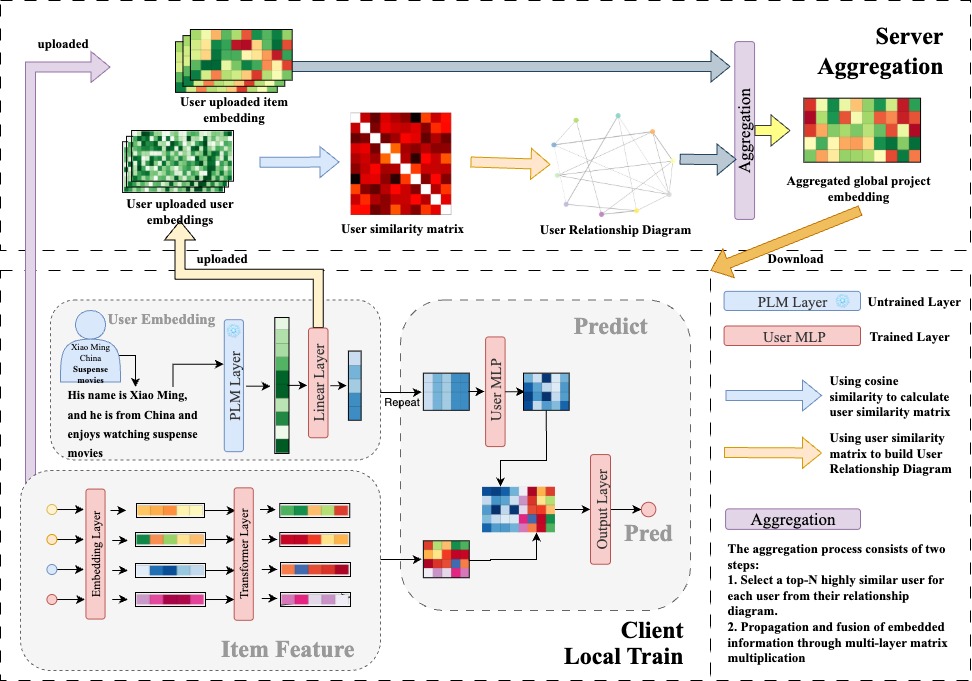}
	\caption{This figure shows a recommendation system framework that integrates federated learning and graph neural networks. Local training of users and item embeddings on the client side, uploading model parameters to the server; The server constructs a relationship graph based on user similarity, aggregates and generates global item embeddings, and sends them back to the client to achieve collaborative recommendation under privacy protection.}
	\label{photo}
\end{figure*}

In the following sections, we will provide a detailed introduction to the proposed UFGraphFR, as illustrated in Figure \ref{photo}.

\subsection{Local Training} 
The client recommendation model consists of five major modules: 

1) \textbf{Joint embedding layer} (module parameter $\theta_{user}$) maps user structured attributes to vector space through natural language prompts templates, using a combination of a pre-trained language model initialization and a trainable linear layer; 

2)  \textbf{Item embedding layer} (module parameter $\theta_{item}$) encodes user interaction item IDs as dense vectors; 

3) \textbf{Temporal Transformer module} (module parameter $\theta_{Transformer}$) models long-term interaction sequences through the mechanism of multi-head self-attention dependencies; 

4) \textbf{User Feature Refinement MLP} (module parameter is $\theta_{\text{umlp}}$) extracts higher-order user representations through a three-layer fully-connected network;  

5) \textbf{Predictive Scoring Function} (module parameter is $\theta_{\text{score}}$) computes user-item embedding splicing based on final prediction results.

\subsubsection{\textbf{Joint Embedding Layer}}
We propose a joint embedding layer to deal with the user's structured features, which contains a three-layer structure. These are the Prompts Output layer, the untrainable pre-trained language model (PLM) layer, and the trainable output layer.

\textbf{Prompt Output:} Each $u$ has a structured data $\mathcal{P}_u$. The prompt output is used to convert the structured data into text language. Taking the 'movielens' dataset as an example, the movielens user data has the following five fields: uid, gender, age, occupation, and zipcode. We can use the following prompt to convert it into natural language: "The user'id is \{\} and his gender is \{\}, he is \{\} years old, works in the field of \{\} and lives at zip code \{\}".

\textbf{PLM Layer: } After converting the structured features $\mathcal{P}_u)$ into natural language cues via $\Gamma$, we obtain a string of user text features. This can in turn be transformed into d-dimensional high-dimensional vectors by freezing a pre-trained language model

\begin{equation} 
	v_u = \text{LM}(\Gamma(\mathcal{P}_u)) \label{con:nlp} 
\end{equation} 
where $\Gamma$ denotes the work done by the Prompts Output Layer, i.e., transforming structured user attributes into natural language cue words, and $LM$ denotes the transformation of the prompts into embedding vectors using a pre-trained language model.  In our method, since the user's structured features $\mathcal{P}_u$ are constant during training and the PLM layer is not trainable, the user's $v_u$ is constant. 

However, since the $\mathcal{P}_u$ are constant during training and the PLM layer is not trainable, the $v_u$ is also constant. This results in a constant user relationship graph constructed from the direct embedding vectors of the uploaded user's text description, and the embedding dimensions of each PLM (e.g., the embedding dimension of the USE model is 100, the embedding dimension of MiniLM-L6 is 384, and that of T5, TinyBERT, and LaBSE are all 768) are not the same. To address these two issues, we add a trainable linear layer after the PLM layer, projecting the high-dimensional, static semantic vector $v_u$ onto a unified, low-dimensional, learnable user preference representation space, as shown in the formula \eqref{con:layers}.

\begin{equation}
	e_u = h(v_u) =  v_u W_{d_1 \times d} + b \label{con:layers}
\end{equation}

Where $W_{i}^{(t)} \in \mathbb{R}^{d_1 \times d}$ is the local personalized weight matrix of the $i$-th user at communication round $t$, and $b$ is the bias term. The key is that each user maintains their own unique weight matrix $W_i$, which is trained and updated using local interaction data, thereby achieving personalized correction of the general semantic vector $v_u$ and generating the final user embedding $e_u$.

According to our training objective (as shown in \eqref{con:fr}), our goal is to train optimal parameters, including the parameters $W_u$ of the joint embedding layer. When the model converges, $W_u$ should accurately reflect user preferences. We choose $W_u$ over $v_u$ primarily because $W_u$ is dynamic and personalized. $v_u$ is a fixed vector encoded by a pre-trained language model (PLM) of the user's static semantic attributes. $W_u$, on the other hand, is the weight matrix of the local linear layer, which encodes how the user's static semantics $v_u$ adapt to their dynamic behavior $\mathcal{D}_u$. It encompasses both the semantics of $v_u$ and reflects the preference patterns of $\mathcal{D}_u$, representing a comprehensive representation of both.

\subsubsection{\textbf{Item Embedding Layer}}
\label{sec:item-embedding-layer}
The Item Embedding Layer constitutes the foundational component of the local recommendation model, with its core being a globally shared, learnable weight matrix \(E_{item} \in \mathbb{R}^{|I| \times d_{item}}\), where \(|I|\) denotes the total size of the global item set and \(d_{item}\) represents the embedding dimension.
During federated training, the server maintains and broadcasts this weight matrix \(\theta_{\text{global}} = E_{item}\) as the globally shared parameter.

This layer performs a deterministic mapping: it converts discrete item identifiers from a user's local interaction history into continuous, low-dimensional vector representations via a lookup table operation.
Formally, given the client's local interaction sequence \(S = \{i_1, i_2, \dots, i_L\}\), the embedding layer retrieves the representation vector for each item in the sequence by indexing the global weight matrix \(E_{item}\):

\begin{equation}
e_{i_t} = E_{item}[i_t, :] \quad \in \mathbb{R}^{d_{item}}, \quad 1 \le t \le L.
\label{eq:item_embedding}
\end{equation}

where \(i_t\) is the ID of the \(t\)-th item in the sequence, \(L\) is the sequence length, and \(e_{i_t}\) is the corresponding dense vector.
These embedding vectors serve as input to subsequent model layers (e.g., the Transformer module).

At the end of each round of communication in federated training, each client $c$ participates in parameter aggregation by uploading its trained parameters ($\theta_{\text{Item}}$) to the server.

\subsubsection{\textbf{Temporal Transformer Module}}\label{sec:transformer-block}
The Transformer Block in the local client model is a core component for capturing temporal dependencies in user-item interaction sequences. It consists of the following key operations:

\textbf{Input Representation} Given a user's historical interaction sequence containing $L$ items, each item $i_t$ has been encoded into a dense embedding vector $e_{i_t}$ by the local Item Embedding Layer $E_{\text{item}}$, as defined in Equation \eqref{eq:item_embedding}. These embeddings are concatenated into a sequence matrix $X \in \mathbb{R}^{L \times d_{\text{item}}}$, which serves as the input to the Transformer Block.

\textbf{Multi-Head Self-Attention Mechanism}
To capture the complex dependencies within user interaction sequences, we adopt a multi-head self-attention mechanism. Specifically, for each element embedded in the sequence, we compute contextualized representations by projecting the input matrix $X \in \mathbb{R}^{L \times d_{\text{item}}}$ into the query($Q$), key($K$) and value($V$) spaces through trainable linear transformations:

\begin{equation}
	Q = XW^Q \quad (W^Q  \in \mathbb{R}^{d_{\text{item}} \times d_h})
\end{equation}
\begin{equation}
	K = XW^K \quad (W^K \in \mathbb{R}^{d_{\text{item}} \times d_h})
\end{equation}
\begin{equation}
	V = XW^V \quad (W^V \in \mathbb{R}^{d_{\text{item}} \times d_h})
\end{equation}

Then, for each attention head $k \in \{1, 2, ..., K_h\}$, we compute the scaled dot-product attention as:

\begin{equation}
	Z = \text{softmax} \left( \frac{QK^T}{\sqrt{d_k}} \right)V \label{con:selfattation}
\end{equation}

Here, $d_h = d_{\text{item}} / K_h$ ensures dimension alignment across multiple heads. This mechanism enables the model to attend to different parts of the sequence simultaneously, enriching the representation of long-range user preferences.

\textbf{Feed-Forward Network}
To enhance the expressiveness of each position-wise representation, we apply a position-wise feed-forward network (FFN) after the self-attention layer. The FFN consists of two linear transformations with a ReLU activation in between:

\begin{equation}
	FFN(X) = \max(0, XW_1 + b_1)W_2 + b_2 \label{con:layer1}
\end{equation}

This module refines the attended representations by introducing non-linearity and higher-level feature interactions, contributing to a more precise modeling of user-item interaction sequences.
\subsubsection{\textbf{User Feature Refinement MLP}}
The User Feature Refinement MLP (UserMLP) is a key module in client-side recommendation models, with its parameters denoted as $\theta_{\text{umlp}}$. Located after the joint embedding layer, this module performs nonlinear transformations and feature enhancements on the low-dimensional initial user embeddings $e_u \in \mathbb{R}^d$ generated by the joint embedding layer to extract more abstract and discriminative high-level semantic representations of users, thereby more accurately characterizing user preferences.

Specifically, this module adopts a three-layer fully connected network structure, but before the user embedding is fed into this MLP, a necessary preprocessing step is performed to align its dimension with the sequential context information from the Temporal Transformer module. As shown in Figure \ref{photo}, the refined user embedding $e_u$ is first repeated $L$ times (where $L$ is the length of the user's historical interaction sequence) to form a matrix $H_u^{(0)} \in \mathbb{R}^{L \times d}$, i.e., $H_u^{(0)} = \text{Repeat}(e_u, L)$. This ensures that each position in the interaction sequence has access to the same, refined user profile information for subsequent fusion. The computation process of the UserMLP on this repeated matrix can then be formally represented as:
\begin{equation}
H_u^{(\ell)} = \sigma \left( H_u^{(\ell-1)} W^{(\ell)} + b^{(\ell)} \right), \quad \ell = 1, 2, 3
\label{eq:user-mlp}
\end{equation}
Where $\ell$ represents the layer index, $H_u^{(0)} \in \mathbb{R}^{L \times d}$ is the input (the repeated user embedding matrix), $W^{(\ell)} \in \mathbb{R}^{d^{(\ell-1)} \times d^{(\ell)}}$ and $b^{(\ell)} \in \mathbb{R}^{d^{(\ell)}}$ are respectively the learnable weights and biases of layer $\ell$. $d^{(\ell)}$ represents the hidden dimension of layer $\ell$. This work adopts a "bottleneck" architecture, i.e., $d^{(1)}=d^{(3)}=d$ and $d^{(2)}=2d$ , to control model complexity while improving representational power. $\sigma(\cdot)$ is the activation function, and ReLU is used here to introduce nonlinearity.
After processing by this three-layer MLP, we obtain the refined user representation matrix:
\begin{equation}
H_u' = H_u^{(3)} \in \mathbb{R}^{L \times d}
\label{eq:refined-user-matrix}
\end{equation}
Each row $h_{u, t}' \in \mathbb{R}^d$ of $H_u'$ corresponds to the refined user representation for the $t$-th position in the interaction sequence. This representation integrates semantic information driven by user text features and is refined through multiple nonlinear transformations. It will serve as one of the key inputs to the prediction scoring function, working together with the sequential context features from the item side to generate the final recommendation for each position.
In the $t$-th round of federated training, the UserMLP parameter of client $i$ is denoted as $\theta_{\text{umlp},i}^{(t)}$. This parameter is a local private parameter, trained and updated only locally on the client using the loss function given in \eqref{con:bce}, without needing to be uploaded to the server. This further enhances the privacy protection capability of the model while protecting users' personalized preferences.
\subsubsection{\textbf{Output Layer}}
As shown in Figure \ref{photo}, the input to this module receives two parts of information:
\begin{itemize}
    \item \textbf{Refined User Embedding \(H_u'\)}: from the User Feature Refinement MLP, representing the user's static profile and higher-order preferences.
    \item \textbf{Sequence Context Representation}: From the latent state of the last position in the Transformer Block's encoded output of the user's historical interaction sequence \(\{e_{i_1}, \ldots, e_{i_L}\}\), reflecting the user's dynamic interests and current behavioral context.
\end{itemize}
To generate the predicted score, we first concatenate these two representations:
\begin{equation}
\text{concat}_i = [H_u' \ || \ \text{Transformer}(X_u)_{\text{final}} ] \in \mathbb{R}^{2d}
\label{eq:concat-input}
\end{equation}
where \(||\) denotes the vector concatenation operation, and \(\text{Transformer}(X)_{\text{final}}\) is the final context representation output by the Transformer module.

Subsequently, the concatenated vector is input into a Multilayer Perceptron (MLP) for nonlinear fusion and score prediction. This MLP is structured as a **multi-layer “tower”** , with dimensions \(2d \to d \to 1\). The calculation process is as follows (using a three-layer MLP as an example):
\begin{align}
z_1 &= \sigma(\text{concat}_i \cdot W_1^{\text{score}} + b_1^{\text{score}}) \\
z_2 &= \sigma(z_1 \cdot W_2^{\text{score}} + b_2^{\text{score}}) \\
\hat{y}_{ui} &= \text{Sigmoid}(z_2 \cdot W_3^{\text{score}} + b_3^{\text{score}})
\label{eq:score-mlp}
\end{align}
where \(W_1^{\text{score}}, W_2^{\text{score}}, W_3^{\text{score}}\) and \(b_1^{\text{score}}, b_2^{\text{score}}, b_3^{\text{score}}\) are the learnable parameters of the MLP, and \(\sigma\) is the ReLU activation function (excluding the last layer Sigmoid). Ultimately, \(\hat{y}_{ui} \in (0, 1)\) represents the predicted preference probability of \(u\) for \(i\), used for supervised learning with the real interaction label \(y_{ui} \in \{0,1\}\) using the BCE loss.
\subsubsection{\textbf{Model Loss Functions}}
For the sake of generality, we discussed a typical scenario that relies only on implicit user-item interaction data for recommendations, i.e. if user $i$ interacts with item $j$, then $r_{ij} = 1$; Otherwise, $r_{ij} = 0$. No primitive features for secondary users (projects) are available. In this model, our loss function consists of two parts. The first part is the binary cross entropy Loss (BCE Loss) of prediction score and label, as shown in formula \eqref{con:bce}. 
\begin{equation} 
	\mathcal{L}_{1}(y,\hat{y})=-(ylog(\hat{y})+(1-y)log(1-\hat{y})) \label{con:bce}
\end{equation}
where $y$ represents the user's label, and $\hat{y}$ represents the predicted score output by the model.
The second part is the regular term of the global item embedding and the user-specific item(sampling negative term) embedding, as shown in formula \eqref{con:re}.
\begin{equation}
	\mathcal{R}(e_{global},e_i^{-}) = Mean((e_{global}-e_{i}^{-})^2) \label{con:re}
\end{equation}
where $e_{i}^{-}$ represents the embedding of sampling negative terms on the client. $e_{global}$  is the global embedded weight of the server-side aggregation.
Based on the above formula, our total loss function is shown in formula \eqref{con:sum}. 
\begin{equation}
	\mathcal{L}_{all}  = 	\mathcal{L}_{1}(y,\hat{y}) + \lambda \mathcal{R}(e_{global},e_i^{-}) \label{con:sum}
\end{equation}
where $\lambda $ is a hyperparameter,  represents the regularization coefficient.. 

\subsection{Federated Training}

In the federated learning scenario, which is shown in Fig. \ref{photo}, where we classify the parameters of the user model into two categories, and the model parameters are classified into two categories based on privacy sensitivity: globally shared parameters $\theta_{global}$ and local private parameters $\theta_{local}$. At each time, the user only uploads the globally shared parameters to the server for parameter aggregation. In our model, global parameters have joint embedding $\theta_{user}$ and
item embedding $\theta_{item}$.

By alternately optimizing the local loss function and the global regular term Eq. \eqref{con:yhmb}, the model achieves cross-client knowledge migration while protecting the privacy of the original interaction data.

\begin{equation}
	\min _{\{ \theta_{all,1} \dots  \theta_{all,N}  \}}  \sum_{i=1}^{N} \mathcal{L}_i( \theta_{all,i}) 
	+ \lambda\mathcal{R}_i ( \theta_{global},  \theta_{item,i} ) \label{con:yhmb}
\end{equation}
where $\theta_{all,i} = \{ \theta_{item,i} , \theta_{user,i} ,\theta_{umlp,i} , \theta_{score,i} , \theta_{Transformer,i}     \}$ is the recommended model parameter for the $i$ th client, and ri is the global user item embed weight aggregated on the server. $ \mathcal{R}(·,·)$  is a regularization term used to constrain local item embedding similar to global user item embedding weights, where $ \lambda$ is the regularization coefficient.

\subsubsection{\textbf{Build User Relationship Graph}}
Traditional federated learning frameworks (e.g., FedAvg\cite{mcmahan_communication-efficient_2023}) use a parameter averaging aggregation strategy:
\begin{equation}
	\theta_{\text{global}}^{t+1} = \frac{1}{N} \sum_{i=1}^N \theta_{\text{item},i}^t
\end{equation}
This method treats all clients equally, but ignores group similarity in user preferences. Therefore, in our model, in order to further capture the correlation between users, we construct a dynamic relationship graph $G=(V, E)$ based on the joint embedding weights uploaded by clients on the server side, which follows the following process:

\textbf{Step1: User Embedding Vectorization}
As described in the local model, the joint embedding layer contains a trainable linear projection layer with a weight matrix $W_i \in \mathbb{R}^{d_1 \times d_2}$, where $d_1$ is the dimension of the text vector generated by the pre-trained language model (PLM), and $d_2$ is the dimension of the final low-dimensional user embedding. To measure semantic similarity between users, we flatten the weight matrix $W_i$ into a vector:
\begin{equation}
w_i = \operatorname{vec}(W_i) \in \mathbb{R}^{d_1 d_2}
\end{equation}

\textbf{Step2: Similarity computation constructed graph}
We use cosine similarity to calculate the semantic similarity between users $i$ and $j$:
\begin{equation}
\mathcal{S}_{ij}  = \frac{w_i \cdot w_j^T}{||w_i||  ||w_j||} \label{con:f1}
\end{equation}
This yields a full user similarity matrix $\mathbf{S} \in \mathbb{R}^{N \times N}$. For each user $i$, we select the $K$ users with the highest similarity (usually $K \ll N$) as its neighbors, thus constructing a sparse, undirected user relationship graph. Let $\mathcal{N}(i)$ be the set of neighbors of user $i$. The adjacency matrix $\mathbf{A}$ of the graph can be defined as:
\begin{equation}
A_{ij} =
\begin{cases}
1, & \text{if } j \in \mathcal{N}(i) \text{ or } i \in \mathcal{N}(j) \\
0, & \text{otherwise}
\end{cases}
\end{equation}
The corresponding degree matrix $\mathbf{D}$ is a diagonal matrix with diagonal elements $D_{ii} = \sum_j A_{ij}$. For the stability of subsequent graph convolution operations, we typically use a symmetrically normalized adjacency matrix:
\begin{equation}
\tilde{\mathbf{A}} = \mathbf{D}^{-\frac{1}{2}} \mathbf{A} \mathbf{D}^{-\frac{1}{2}}
\end{equation}

\subsubsection{\textbf{Server-side Global Item Embedding Learning}}
Based on the constructed user relationship graph, we employ a lightweight Graph Convolutional Network (GCN) to aggregate neighbor information on the server side and update the global item embeddings. Specifically, let $\mathbf{E}^{(t)} \in \mathbb{R}^{N \times d_{item}}$ represent the matrix of local item embeddings uploaded by all clients in the $t$-th communication round (where the $i$-th row $\mathbf{e}_i^{(t)}$ corresponds to the local item embedding vector of user $i$). We perform a single-layer graph convolution operation to aggregate neighbor information:
\begin{equation}
\mathbf{R}^{(t)} = \tilde{\mathbf{A}} \mathbf{E}^{(t)} \label{con:gcn_operation}
\end{equation}
where $\mathbf{R}^{(t)} \in \mathbb{R}^{N \times d_{item}}$ is the intermediate representation matrix after graph convolution, incorporating neighbor information.

\subsubsection{\textbf{Parameter Update and Broadcasting}}
The server obtains the new global item embedding by averaging each row of $\mathbf{R}^{(t)}$ (i.e., the item embedding corresponding to each user) along the user dimension:
\begin{equation}
\theta_{global}^{(t+1)} = \frac{1}{N} \sum_{i=1}^{N} \mathbf{r}_i^{(t)} \label{con:global_update}
\end{equation}
Here, $\mathbf{r}_i^{(t)}$ is the $i$-th row of $\mathbf{R}^{(t)}$. Finally, the server broadcasts the updated $\theta_{global}^{(t+1)}$ to all clients for the next round of local model initialization. This graph-based aggregation mechanism allows users with similar textual features (reflected through the joint embedding layer weights) to share item embedding knowledge, thereby enhancing the model's collaborative filtering capabilities while strictly protecting the privacy of users' original interaction data.

\subsubsection{\textbf{The effectiveness of choosing $W_u$}}
In UFGraphFR, the core reason for choosing a user-specific weight matrix \(W_u\) instead of a static semantic vector \(v_u\) to construct the user relationship graph is that \(W_u\) is a dynamic and personalized comprehensive representation. Specifically, the static semantic vector \(v_u\) is generated by a fixed pre-trained language model and only represents the user's inherent attributes (such as age and occupation), failing to reflect their dynamic preferences; while the weight matrix \(W_u\) is trained using local interaction data \(D_u\), and its update process (\(W_u^{(t+1)} = W_u^{(t)} - \eta \nabla_{W_u} \mathcal{L}_{\text{all}}\)) allows it to adapt the static semantics \(v_u\) to the dynamic behavior \(D_u\).

When training converges to the optimal solution, \(W_u\) reaches a steady-state value \(W_u^*\), which simultaneously encodes two key pieces of information: on the one hand, since \(e_u = v_u W_u^* + b\) depends on \(v_u\), \(W_u^*\) retains the user's static semantics; on the other hand, because the update of \(W_u^*\) is achieved by minimizing the loss function based on \(D_u\), it also captures the user's dynamic behavior patterns. Therefore, \(W_u^*\) can be formally represented as \(W_u^* \approx F(v_u, D_u)\), a mapping function that fuses static attributes with dynamic preferences.

This fusion characteristic allows the user relationship graph constructed based on \(W_u\) to more accurately depict the similarity of preferences between users. In contrast, a graph constructed directly using \(v_u\) is only determined by static attributes and cannot reflect behavioral similarity. Therefore, in UFGraphFR, the server uses \(W_u\) to calculate user similarity and construct the relationship graph, enabling subsequent global model aggregation (such as \(\theta_{\text{global}}^{(t+1)} = \frac{1}{N} \sum_{i=1}^{N} \mathbf{r}_i^{(t)}\)) to achieve knowledge sharing among similar users, thereby significantly enhancing collaborative filtering capabilities while strictly protecting privacy.

\subsection{Algorithm}
Based on the above, after several rounds of iteration, we can optimize the whole model parameters. The algorithm we propose is shown in Algorithm \ref{alg:ufgraphfr}. The whole idea of our algorithm is to use the text feature description of the user to build the user's relationship graph, and use the relationship graph and federated parameter aggregation method to learn the global item embedding that can reflect the universality.

\begin{breakablealgorithm}
\caption{UFGraphFR’s algorithm}\label{alg:ufgraphfr} 

\begin{algorithmic}[1]
\REQUIRE Number of clients $N$, total communication rounds $T$, local epochs $E$, learning rate $\eta$, regularization coefficient $\lambda$, user joint-embedding layer parameters $\theta_{user}$, item embedding layer parameters $\theta_{item}$, and other local model parameters $\theta_{local}$.
\ENSURE Optimized global item embedding $\theta_{global}$.
\STATE \textbf{Initialize} global parameters $\theta_{global}^{(0)}$ (e.g., item embeddings) on the server.
\STATE \textbf{Broadcast} $\theta_{global}^{(0)}$ to all $N$ clients.
\FOR{each communication round $t = 1, 2, ..., T$}
    \STATE \textbf{Client Local Training (Parallel Execution):}
    \FOR{each client $i = 1, 2, ..., N$}
        \STATE Initialize local item embedding $\theta_{item,i}^{(t)} \leftarrow \theta_{global}^{(t-1)}$.
        \STATE Use local data to train the local model for $E$ epochs by minimizing the loss function $\mathcal{L}_{all}$ (Eq.~\eqref{con:sum}) via SGD.
        \STATE Obtain updated local parameters: $\{\theta_{item,i}^{(t)}, \theta_{user,i}^{(t)}, \theta_{Transformer,i}^{(t)}, \theta_{umlp,i}^{(t)}, \theta_{score,i}^{(t)}\}$.
        \STATE \textbf{Upload} the user joint-embedding weight matrix $W_i^{(t)} \in \mathbb{R}^{d_1 \times d_2}$ and the local item embedding $\theta_{item,i}^{(t)}$ to the server.
    \ENDFOR
    \STATE \textbf{Server Update with Graph Aggregation:}
\STATE \textbf{Vectorize:} Flatten each client’s uploaded joint-embedding weights:  
$w_i^{(t)} = \text{vec}(W_i^{(t)}) \in \mathbb{R}^{d_1 d_2}$.
\STATE \textbf{Compute Similarity:} Calculate cosine similarities $\mathcal{S}_{ij}^{(t)}$ between all $w_i^{(t)}$ and $w_j^{(t)}$.
\STATE \textbf{Build Graph:} Construct user graph $\mathcal{G}^{(t)}$ by connecting each user to its top-$k$ similar users. Obtain adjacency matrix $\mathcal{A}^{(t)}$ and degree matrix $\mathcal{D}^{(t)}$.
\STATE \textbf{Aggregate Embeddings:} Apply GCN to local item embeddings $A^{(t)}$:  
$\mathcal{R}^{(t)} = (\mathcal{D}^{(t)})^{-1/2} \mathcal{A}^{(t)} (\mathcal{D}^{(t)})^{-1/2} A^{(t)}$.
\STATE \textbf{Update Global Embedding:} Set $\theta_{\text{global}}^{(t)} = \frac{1}{N} \sum_{i=1}^N \mathcal{R}_i^{(t)}$.
\STATE \textbf{Broadcast} $\theta_{\text{global}}^{(t)}$ to all clients.
\ENDFOR
\RETURN The final global item embedding $\theta_{global}^{(T)}$.
\end{algorithmic}
\end{breakablealgorithm}

\subsubsection{Improve client efficiency}In real-world scenarios, user feature attributes and items in recommender systems are often very large, which poses potential embedded storage and communication overhead challenges for resource-limited client devices. To address this problem, we propose that each user is embedded with the user's textual features only in the first round and when the user's information changes.

\subsection{Privacy Protection}

Under the federated learning framework, our approach inherits the privacy benefit of keeping raw data locally on each user's device. However, the process of uploading model parameters after each training round presents potential privacy risks. To mitigate this, we incorporate local differential privacy (LDP) \cite{choi_guaranteeing_2018} by adding Laplacian noise (as shown in Formula \eqref{con:f2}) to the parameters before uploading them. This mechanism provides an additional layer of privacy protection by introducing uncertainty into the client-side updates.

\begin{equation}
    \theta_{i} = \theta_{i} + \text{Laplace}(0, \epsilon)
    \label{con:f2}
\end{equation}
where $\alpha$ controls the noise scale. A larger $\epsilon$ introduces more noise, offering stronger privacy guarantees but potentially affecting model utility. It is crucial to note that in this work, we treat this mechanism primarily as a tunable noise regularizer or a heuristic privacy-enhancing technique. Our empirical analysis  shows that carefully chosen noise levels can maintain model utility while obfuscating parameters. However, we do not claim formal local differential privacy (LDP) guarantees because we have not conducted the necessary formal analysis (e.g., bounding the sensitivity of the upload function or accounting for privacy composition across multiple training rounds). Future work will focus on integrating this noise addition into a rigorous LDP or DP-SGD framework to provide certified guarantees.

\section{Experiment}
This section analyzes the proposed methods through experiments, aiming to answer the following questions:
\begin{itemize}
	\item  \textbf{Q1:} Is UFGraphFR superior to current advanced federated and centralized recommendation models?
	\item  \textbf{Q2:} Do the modules we introduce make the unintroduced modules perform better?
	\item \textbf{Q3:} How will UFGraphFR perform with different pre-trained language models?
	\item  \textbf{Q4:} Does the noise introduced by UFGraphFR in increasing local segmentation privacy affect performance?
\end{itemize}
\subsection{Datasets and Evaluation Protocols}

\subsubsection{\textbf{datasets}}We validated the proposed UFGraphFR on four recommended benchmark datasets: MovieLens-100K, MovieLens-1M\cite{harper_movielens_2016}, Lastfm-2K\cite{cantador_second_2011} and HetRec2011\cite{cantador_second_2011}. In particular, two MovieLens datasets were collected from the MovieLens website, recording user ratings for movies, with no fewer than 20 ratings per user. Lastfm-2K is a music dataset where each user keeps a list of artists listened to and a listen count. We removed users with less than 5 interactions from Lastfm-2K. HetRec2011 is an extension of MovieLens-10M, which connects movies with the corresponding web pages of the Internet Movie Database (IMDb) and the Rotten Tomatoes movie review system. For the Lastfm-2K and HetRec2011 datasets, we used the user id and the total number of user interaction items as user attributes. Detail statistics are shown in Table.\ref{tab:dataset}   
\begin{center}
    \begin{table}[!ht]
	\caption{Dataset statistics}
	\label{tab:dataset}
	\centering
	\begin{tabular*}{\linewidth}{p{3cm} p{3cm} p{3cm} p{3cm} p{3cm}}
		\hline
		Name & Users & Items & Interactions & Sparsity\\
		
		\hline
		MovieLens-100K & 943 & 1682&  100,000 & 93.70\%
		\\
		MovieLens-1M & 6,040 &3,706 &1,000,209 &95.53\%
		\\
		Lastfm-2K & 1,600 & 12,454&  185,650 &  99.07\% \\
		HetRec2011 & 2,113 & 10,109 & 855,598 &  95.99\% \\  \hline

	\end{tabular*}
	
\end{table}
\end{center}
\subsubsection{\textbf{Evaluation protocols.}}

To ensure the fairness and reproducibility of the evaluation results, and to fully reflect the advantages of the Transformer module used in the model for modeling temporal dependencies, we adopted a strict within-user leave-one-out method based on time series, using hit rate (HR@K) and normalized depreciation cumulative gain (NDCG@K) as the core evaluation metrics. The specific data processing and evaluation process is as follows:

\textbf{Dataset Partitioning}: For a dataset containing interaction records (user ID, item ID, timestamp): For each user, all their interaction records are first sorted in ascending order by timestamp. The latest (last) interaction in the user's sequence is used as the test item. The second-to-last (nearest) interaction in the user's sequence is used as the validation item. All remaining earlier interaction records for that user constitute the training set. This strategy of "retaining only the last two interactions as validation and testing" aligns with the real-world setting of time-series recommendation scenarios, i.e., using past user behavior (training set) to predict future behavior (validation/test set).

\textbf{Candidate Set Construction and Model Evaluation}: To efficiently evaluate the model's Top-K recommendation performance, we employ a global sampling-based evaluation protocol: During the evaluation phase (both on the validation and test sets), for the target positive item (ground-truth item) corresponding to user u, we do not restrict the candidate set of interactions before the test interaction to the user's uninterrupted item pool. Instead, to simulate the scenario of a real recommendation system with a large item library, we randomly sample 99 "negative" items for the user from the full item pool. These 99 items may overlap with items in the user's training set or historical interactions, but they definitely do not include the user's test item itself. These 1 positive item and 99 random negative items together constitute an evaluation candidate item pool of size 100. The model independently predicts the probability that the user will be interested in all 100 items in this candidate pool and generates a ranked list. Finally, the HR@K and NDCG@K metrics are calculated based on the position of the positive item in the ranked list. For example, if the positive item appears in the Top-10 list, then HR@10 is counted as a hit.
\subsection{Baselines and Implementation Details}
\subsubsection{Baselines}
We compared our approach to two baseline branches, including a centralized and federated recommendation model. All methods make recommendations based solely on user-project interactions.
\begin{itemize}
	
	\item  \textbf{Matrix Factorization (MF)}\cite{mf} A typical recommendation model. It decomposes the scoring matrix into two embeddings in the same hidden space, describing user and item characteristics, respectively.
	\item  \textbf{Neural Collaborative Filtering (NCF)}\cite{ncf} This method is one of the most representative neural recommendation models. We first learn the user-embedded module and the item-embedded module, and then model the user's item interaction by MLP.
	\item  \textbf{Self-supervised Graph Learning (SGL)}\cite{sfg} This method is a self-supervised graph learning enhanced recommendation model.
	
	\item  \textbf{FedMF}\cite{chai_secure_2021} FedMF trains users locally to embed and uploads project gradients to the server for global aggregation.
	\item  \textbf{FedNCF}\cite{10.1016/j.knosys.2022.108441} FedNCF treats user embeddings as private components of local training and shares project embeddings and MLPS to perform collaborative training.
	\item  \textbf{Federated Reconstruction (FedRecon)}\cite{singhal_federated_2022} An advanced personalized federated learning framework, FedRecon retrains user embeddings in each round and computes item gradients based on the retrained user embeddings.
	\item  \textbf{Meta Matrix Factorization (MetaMF)}\cite{lin_meta_2020} It is a distributed matrix decomposition framework in which meta-networks are used to generate fractional functional modules and private term embeddings.
	\item  \textbf{Personalized Federated Recommendation (PFedRec)}\cite{zhang2023dual} It is a personalized federated recommendation framework where the server first learns a common project embed for all clients and then fine-tunes the project embed with local data for each client.
	\item  \textbf{Federated LightGCN (FedLightGCN)}\cite{zhang_gpfedrec_2024} The way LightGCN\cite{he_lightgcn_2020} extends to the federated learning framework in the GPFedRec paper. In particular, each client trains local LightGCN using a first-order interaction subgraph.
	\item  \textbf{Federated Graph Neural Network
    (FedPerGNN)} \cite{wu_federated_2022} It deploys a graph neural network on each client, and users can integrate high-level user-project information through a graph extension protocol.
	
	\item  \textbf{Graph-Guided Personalization for Federated Recommendation (GPFedRec)}\cite{zhang_gpfedrec_2024} GPFedRec is a graph-guided federated recommendation system, which proposes a graph-guided aggregation mechanism. Our model is mainly referenced from this model.
\end{itemize}

\subsubsection{\textbf{Implementation details}}

We implemented the UFGraphFR method based on the PyTorch framework and developed its lightweight variant, UFGraphFR-Lite, which improves efficiency by periodically updating the user relationship graph, avoiding the need for the client to upload user embeddings in each round of communication. All experiments were conducted on multiple datasets (including lastfm-2k, ml-1m, and 100k) and followed a uniform experimental setup to ensure fair comparison. We set the embedding dimension to 32, used a fixed batch size of 256, and set the total training epochs (for centralized methods) or communication epochs (for federated methods) to 100 to ensure convergence for all methods; for federated learning, the number of local training epochs per round was set to 1. The score function module in the model adopts a three-layer MLP architecture with a structure of 64→32→8→1. For the text embedding part, we loaded a USE\cite{cer_universal_2018} pre-trained model using MediaPipe\cite{lugaresi2019mediapipeframeworkbuildingperception}, setting the text embedding dimension to 100, while the output dimension of the joint embedding layer was 32. To better capture user details, the user-side MLP layer adopts a two-layer architecture of 32-64-32, consistent with the feedforward network structure in our Transformer module. Furthermore, we configured all hyperparameters in detail, including similarity metric, neighbor selection strategy, number of negative samples, learning rate, regularization coefficient, etc., and set up a complete random seed mechanism to ensure the reproducibility of the experiments. All experimental parameter configurations are provided in Appendix \ref{sec:appendix} of the supplementary materials.

\begin{table}
	\caption{Performance comparison on four datasets. }
	\label{perfer}
	\begin{tabular*}{\linewidth}{p{0.5cm} c cccccccc}
			\hline
			\multicolumn{1}{l}{\multirow{2}{*}{}} & \multirow{2}{*}{Method} & \multicolumn{2}{c}{MovieLens-100K}                      & \multicolumn{2}{c}{MovieLens-1M}                        & \multicolumn{2}{c}{Lastfm-2K}                           & \multicolumn{2}{c}{HetRec2011}                          \\
			\multicolumn{1}{l}{}                  &                         & \multicolumn{1}{c}{HR@10} & \multicolumn{1}{c}{NDCG@10} & \multicolumn{1}{c}{HR@10} & \multicolumn{1}{c}{NDCG@10} & \multicolumn{1}{c}{HR@10} & \multicolumn{1}{c}{NDCG@10} & \multicolumn{1}{c}{HR@10} & \multicolumn{1}{c}{NDCG@10} \\ \hline
			\multirow{3}{*}{CenRec}               & MF                      & 64.58                     & 38.69                       & 68.70                     & 41.47                       & 83.13                     & 71.78                       & 66.07                     & 41.21                       \\
			& NCF                     & 64.21                     & 37.13                       & 64.02                     & 38.16                       & 82.57                     & 68.26                       & 64.74                     & 39.55                       \\
			& SGL                     & 64.9                      & 40.02                       & 62.6                      & 34.13                       & 82.37                     & 68.59                       & 65.12                     & 40.18                       \\ \hline
			\multirow{8}{*}{FedRec}               & FedMF                   & 66.17                     & 38.73                       & 67.91                     & 40.81                       & 81.63                     & 68.18                       & 64.69                     & 40.29                       \\
			& FedNCF                 & 60.66                     & 33.93                       & 60.38                     & 34.13                       & 81.44                     & 61.95                       & 60.86                     & 36.27                       \\
			& FedRecon               & 65.22                     & 38.49                       & 62.78                     & 36.82                       & 82.06                     & 67.37                       & 61.57                     & 34.2                        \\
			& MetaMF                 & 66.21                     & 41.02                       & 44.98                     & 26.31                       & 81.04                     & 64.13                       & 54.52                     & 32.36                       \\
			& PFedRec                & 71.37                     & 42.59                       & \underline{73.03}                     & \underline{44.49}                       & 82.38                     & 73.19                       & 67.2                      & 42.7                        \\
			& FedLightGCN$^{2}$             & 24.53                     & 12.78                       & 37.53                     & 15.01                       & 43.75                     & 15.17                       & 22.65                     & 7.96                        \\
			& FedPerGNN$^{3}$               & 11.52                     & 5.08                        & 9.31                      & 4.09                        & 10.56                     & 4.25                        & –                         & –                           \\
			& GPFedRec                & \underline{72.65}                     &  \underline{43.67}                       & 72.26                     & 43.17                          & \underline{83.44}                     & \underline{74.11}                       & \underline{69.71 }                    & \underline{43.17}                       \\ \hline
			\multirow{2}{*}{Ours}                 & UFGraphFR               & \textbf{76.03*}           & \textbf{47.31*}             & \textbf{75.55*}           & \textbf{46.32*}             & \textbf{85.69*}           & \textbf{77.32*}             &70.42          & \textbf{45.17*}             \\
			& UFGraphFR-Lite          & 75.19                     & 46.46                       & 71.04                     & 42.49                       & 85.19                     & 77.32                       &   \textbf{70.56*}                 & 44.95                       \\ \hline
			\multicolumn{2}{c}{Improvement}                                 & 
            $\uparrow$3.38                      &  $\uparrow$3.64                      &  $\uparrow$2.52                      &  $\uparrow$1.83                        &  $\uparrow$2.25      & 
            $\uparrow$ 3.21                       &  $\uparrow$0.85                     &  $\uparrow$2.00                        \\ \hline             
		\end{tabular*}
	\begin{tablenotes}
\item 1. Performance comparison of the four datasets. The best results are in bold and the best baseline results are underlined. FedPerGNN could not run on HetRec2011 due to unacceptable memory allocation (indicated by '-'). “ to indicate this). "*" and 'Improvement' denote statistically significant improvement and performance improvement over the best baseline, respectively. $\uparrow$ denotes the difference in enhancement relative to the best baseline. 
\item 2. FedLightGCN performs poorly primarily because it can only be trained using local first-order interaction subgraphs. This prevents the server from constructing a complete global user-item interaction graph, severely limiting its ability to capture high-order collaboration signals between users.
\item 3.  FedPerGNN performs poorly mainly because it requires aligning user history interaction records to discover high-order neighbors. This leads to negative sampling bias (matching neighbors with non-interacting negative samples), reducing the accuracy of actual neighbor discovery and introducing high computational overhead, resulting in poor scalability when handling large-scale datasets.
	\end{tablenotes}            
	
\end{table}

\subsection{Performance (Q1)}
Table \ref{perfer} shows the performance of HR and NDCG on four datasets in the Top-10 recommendation scenario. 

\textbf{Our approach achieves better performance than the centralized recommendation system approach in all Settings.} In the Top-10 recommendation scenario, UFGraphFR outperforms existing methods under all settings. Specifically, on the MovieLens-100K dataset, UFGraphFR achieves performance improvements of up to 4.65\% and 8.30\% in HR@10 and NDCG@10, respectively, compared to the best centralized baseline (MF) and federated baseline (GPFedRec). This advantage primarily stems from the following points: First, unlike centralized methods that store all user-item parameters on the server, UFGraphFR treats user embeddings and rating functions as private components on the client side, enabling personalized learning. More importantly, this framework innovatively utilizes user text feature descriptions (i.e., structured attributes encoded by natural language prompts and pre-trained language models) to securely construct a user relationship graph on the server side, and combines it with a lightweight graph neural network for information propagation, effectively mining and utilizing semantic relationships between users. Furthermore, the Transformer module in the local model further enhances the ability to model temporal dependencies in user interaction sequences. Finally, our lightweight variant, UFGraphFR-Lite (as shown in Table 4, which updates the user relationship graph every 5 rounds), achieves comparable or even better performance than the full model on most datasets, striking a good balance between model efficiency and performance.

\subsection{Ablation experiment (Q2)}
Table \ref{tab:ablation_study_all} presents the results of ablation experiments on the UFGraphFR model, aiming to evaluate the contribution of two key modules of the model  the Transformer module and the Joint Embedding module to the recommendation performance under different datasets (MovieLens-100K, MovieLens-1M, Lastfm-2K, HetRec2011). The table reports the performance of each configuration on both HR@10 and NDCG@10 metrics while giving the improvement values compared to the baseline model GPFedRec.

\begin{table}
	\caption{Ablation study results of UFGraphFR on four datasets. GPFedRec is used as the baseline. Performance drops relative to the full model are shown below each variant. }
	\label{tab:ablation_study_all}
	\begin{tabular*}{\linewidth}{cccccccccc}
				\hline
				\multicolumn{1}{l}{\multirow{2}{*}{}} & \multirow{2}{*}{Method} & \multicolumn{2}{c}{MovieLens-100K} & \multicolumn{2}{c}{MovieLens-1M} & \multicolumn{2}{c}{Lastfm-2K} & \multicolumn{2}{c}{HetRec2011} \\
				\multicolumn{1}{l}{} & & HR@10 & NDCG@10 & HR@10 & NDCG@10 & HR@10 & NDCG@10 & HR@10 & NDCG@10 \\ 
				\hline
         
			& \multirow{2}{*}{\textbf{\makecell{GPFedRec} }}
				& \multirow{2}{*}{\centering 72.65} & \multirow{2}{*}{\centering 43.67}
				& \multirow{2}{*}{\centering 72.26} & \multirow{2}{*}{\centering 43.17}
				& \multirow{2}{*}{\centering 83.44} & \multirow{2}{*}{\centering 74.11}
				& \multirow{2}{*}{\centering 69.71} & \multirow{2}{*}{\centering 43.17} \\
				\\
				\hline

				& \makecell{UFGraphFR-nT}        & 72.85 & 44.73 & 72.80 & 43.84 & 83.94 & 75.66 & 69.33 & 43.37 \\
				& ↓ vs Full                           & $\downarrow$3.18 & $\downarrow$2.58 & $\downarrow$2.75 & $\downarrow$2.48 & $\downarrow$1.75 & $\downarrow$1.66 & $\downarrow$1.09 & $\downarrow$1.80 \\
				& \textcolor{red}{$\uparrow$ vs Baseline} & $\uparrow$0.20 & $\uparrow$1.06 & $\uparrow$0.54 & $\uparrow$0.67 & $\uparrow$0.50 & $\uparrow$1.55 & $\downarrow$0.38 & $\uparrow$0.20 \\
				\hline
				& \makecell{UFGraphFR-nJ}      & 74.23 & 46.40 & 74.40 & 45.21 & \textbf{85.94} & \textbf{77.52} & 69.43 & 42.78 \\
				& ↓ vs Full                           & $\downarrow$1.80 & $\downarrow$0.91 & $\downarrow$1.15 & $\downarrow$1.11 & $\uparrow$0.25 & $\uparrow$0.20 & $\downarrow$0.99 & $\downarrow$2.39 \\
				& \textcolor{red}{$\uparrow$ vs Baseline} & $\uparrow$1.58 & $\uparrow$2.73 & $\uparrow$2.14 & $\uparrow$2.04 & $\uparrow$2.50 & $\uparrow$3.41 & $\downarrow$0.28 & $\downarrow$0.39 \\
				\hline
				& \makecell{UFGraphFR-nB}                 & 73.38 & 44.69 & 73.51 & 44.48 & 84.50 & 75.73 & 69.05 & 42.39 \\
				& ↓ vs Full                           & $\downarrow$2.65 & $\downarrow$2.62 & $\downarrow$2.04 & $\downarrow$1.84 & $\downarrow$1.19 & $\downarrow$1.59 & $\downarrow$1.37 & $\downarrow$2.78 \\
				& \textcolor{red}{$\uparrow$ vs Baseline} & $\uparrow$0.73 & $\uparrow$1.02 & $\uparrow$1.25 & $\uparrow$1.31 & $\uparrow$1.06 & $\uparrow$1.62 & $\downarrow$0.66 & $\downarrow$0.78 \\
				\hline
                & \makecell{UFGraphFR-oT}                 & 75.19 & 46.75 & 73.87 & 45.19 & 84.81 & 76.99 & 69.14 & 44.14 \\

				& ↓ vs Full                           & $\downarrow$0.84 & $\downarrow$0.56 & $\downarrow$1.68 & $\downarrow$1.13 & $\downarrow$0.88 & $\downarrow$0.33 & $\downarrow$1.28 & $\downarrow$1.03 \\
				& \textcolor{red}{$\uparrow$ vs Baseline} & $\uparrow$2.54 & $\uparrow$3.08 & $\uparrow$1.61 & $\uparrow$2.02 & $\uparrow$1.37 & $\uparrow$2.88 & $\downarrow$0.57 & $\uparrow$0.97 \\
				\hline
				& \multirow{2}{*}{\textbf{\makecell{UFGraphFR} }}
				& \multirow{2}{*}{\textbf{76.03}} & \multirow{2}{*}{\textbf{47.31}}
				& \multirow{2}{*}{\textbf{75.55}} & \multirow{2}{*}{\textbf{46.32}}
				& \multirow{2}{*}{85.69} & \multirow{2}{*}{77.32}
				& \multirow{2}{*}{\textbf{70.42}} & \multirow{2}{*}{\textbf{45.17}} \\ \\
				\hline

    \end{tabular*}
    \begin{tablenotes}
    \item $\downarrow$ denotes the performance degradation relative to the full UFGraphFR model.
    \item $\uparrow$ denotes the performance improvement relative to the baseline model GPFedRec.
    \item Bold values indicate the highest performance for each dataset.
    \item UFGraphFR-nT (Without Transformer): Version where the Transformer block is removed.
    \item UFGraphFR-nJ (Without Joint Embedding): Version where the Joint Embedding module is replaced by a standard embedding layer.
    \item UFGraphFR-nB (Without Both): Version where both the Transformer and Joint Embedding modules are removed (this variant retains the additional UserMLP layer compared to GPFedRec).
    \item UFGraphFR-oT (original text embedding): Version where the user uploads their original pre-trained model text vector ($v_u$) instead of the learnable linear layer weight ($W_u$).
    \end{tablenotes}
	
\end{table}

In our ablation studies, for UFGraphFR-nT, we remove the Transformer block on the client side. For UFGraphFR-nJ, we replace the Joint Embedding layer with a standard user embedding layer. UFGraphFR-nB implements both modifications. Notably, compared to the baseline GPFedRec \cite{zhang_gpfedrec_2024}, even UFGraphFR-nB retains an additional UserMLP layer for user preference refinement.

\subsubsection{no Transformer}

Removing the Transformer module (UFGraphFR-nT) consistently degrades performance, particularly in terms of NDCG. For example, on MovieLens-1M, NDCG@10 drops from 46.32 (full model) to 43.84. This confirms that the Transformer's ability to model long-term dependencies in user interaction sequences is crucial for achieving accurate rankings.

\subsubsection{no Joint Embedding}

Replacing the Joint Embedding with a standard layer (UFGraphFR-nJ) shows mixed results. While it slightly improves performance on the Lastfm-2K dataset, it leads to stable and significant gains on the MovieLens series and HetRec2011 for both HR and NDCG. This underscores the general utility of incorporating semantic text features for user relationship modeling in federated settings.

\subsubsection{no Both}

The model with both modules removed (UFGraphFR-nB) shows a small performance gain over GPFedRec on most datasets, attributed to the retained UserMLP. However, it suffers a notable performance drop compared to the full UFGraphFR, e.g., a near 3-point decrease in HR@10 on MovieLens-100K. This highlights the complementary role of the Joint Embedding and Transformer modules.

\subsubsection{Original Text Embedding (UFGraphFR-oT)}

The variant UFGraphFR-oT, which uses the static PLM output ($v_u$) instead of the learned linear layer weights ($W_u$) for graph construction, generally underperforms the full model. This demonstrates the advantage of using the dynamic, preference-adapted $W_u$ for building a more accurate and personalized user relationship graph. In the actual experiment, we found that our hypothesis "similar user text feature descriptions have similar preferences" has a relatively weak positive correlation. This is a direction that we need to improve in our future research. How to quantify features better is the future goal.

In summary, the ablation studies validate that the proposed Joint Embedding and Transformer modules are key contributors to UFGraphFR's performance. Their combined use in the full model achieves the optimal balance, leveraging both semantic user features and sequential behavior patterns for accurate and personalized federated recommendations.
\subsection{PLM \& Modeling Effect(Q3)}
In this subsection, we will evaluate the use of different pre-training models to train UFGraphFR. Specifically, we select the following model to train the model on the 100k dataset: 
\begin{itemize}
	\item  \textbf{Universal Sentence Encoder,USE} \cite{cer_universal_2018} it provided by MediaPipe\cite{lugaresi2019mediapipeframeworkbuildingperception} to embed text. The text embedding dimension is 100.
	\item  \textbf{LaBSE}\cite{feng-etal-2022-language} The language-agnostic BERT sentence embedding encodes text into high-dimensional vectors. The Embedding dimension is 768.
	\item \textbf{GTR-T5-Base,T5}\cite{ni2021largedualencodersgeneralizable} This is a sentence transformers model: It maps sentences \& paragraphs to a 768 dimensional dense vector space. 
	\item  \textbf{all-MiniLM-L6-v2,MiniLM} This is a sentence transformers model: It maps sentences \& paragraphs to a 384 dimensional dense vector space and can be used for tasks like clustering or semantic search.
	
	\textbf{TinyBERT-L6,TinyBERT}\cite{reimers-2019-sentence-bert}
	This is a sentence-transformers model: It maps sentences \& paragraphs to a 768 dimensional dense vector space and can be used for tasks like clustering or semantic search.
\end{itemize}

The experimental results are shown in Table \ref{plmmodel_performance}. It  shows the performance comparison of different PLMs on recommendation tasks, with evaluation metrics of HR@10 (Hit Rate) and NDCG@10 (Normalized Discounted Cumulative Gain).

The experimental results indicate that all PLMs are HR@10 and NDCG@10 the two indicators are superior to the baseline model GPFedRec, indicating that the graph structure constructed through user text features has significant advantages in federated recommendation. Among all models, TinyBERT performs the best (HR@10 : 76.35, NDCG@10 : 47.36), demonstrating its powerful semantic modeling ability and competitive advantage in lightweight models.

\begin{table}[ht]
	\centering
	\caption {Performance comparison of PLMs}
	
	\label{plmmodel_performance}
		
		\begin{tabular}{lll}
			\hline
			\textbf{Model}           & \textbf{HR@10} & \textbf{NDCG@10} \\ \hline
			\textbf{GPFedRec}                      & 72.85                               &  43.77                                 \\ 
			UFGraphFR-USE & 75.72 & 47.08 \\ 
			UFGraphFR-MiniLM-L6 & 74.87 & 46.63 \\
			UFGraphFR-T5 & 76.14 & 46.60 \\ 
			UFGraphFR-TinyBERT & \textbf{76.35} & \textbf{47.36} \\
			UFGraphFR-LaBSE & 75.50 & 46.89 \\ \hline
		\end{tabular}
	
\end{table}

\subsection{Privacy Protection (Q4)}
In this subsection, we evaluate the performance of our privacy protection enhanced UFGraphFR with the local differential privacy strategy. Particularly, we set the noise intensity $\epsilon$ = [0.05, 0.1, 0.2, 0.3, 0.4], and experimental results are shown in Table \ref{dps}.
\begin{table}[h]
	\caption{Performance on different $\epsilon$. }
	\label{dps}
	\begin{tabular}{llllll}
				\hline
				& 0.05  & 0.1   & 0.2   & 0.3   & 0.4   \\ \hline
				HR@10   & 75.93 & 75.63 & 74.72 & 74.33 & 73.97 \\
				NDCG@10 & 47.12 & 46.86 & 46.62 & 46.30 & 45.87 \\ \hline
				
			\end{tabular}
	
\end{table}

Based on experimental data at different noise intensities $\epsilon$, we found that the model performance (HR@10 and NDCG@10) exhibits a monotonically decreasing trend with increasing $\epsilon$, but the magnitude of the decrease remains relatively gradual within a reasonable noise introduction range. Specifically, at lower noise levels (e.g., $\epsilon$ = 0.05 and 0.1), HR@10 and NDCG@10 maintain high performance, indicating that moderately introducing noise enhances privacy protection while the reduction in recommendation accuracy is within acceptable limits. As the noise intensity further increases ($\epsilon$ = 0.2 to 0.4), an observable decrease in model performance occurs, suggesting that excessive noise may interfere with the model's ability to capture data patterns. Overall, these results demonstrate that within our architectural framework, a practical balance can be achieved between model utility and enhanced privacy protection by carefully adjusting the noise intensity. The introduction of noise, as an architectural-level mitigation measure, primarily increases the uncertainty of uploaded parameters, thereby improving the actual level of privacy protection, rather than providing quantified differential privacy guarantees. This provides a feasible practical reference for deploying federated recommender systems in privacy-sensitive scenarios.

\subsection{Convergence analysis}
\begin{figure*}[hbpt]
	\centering
	\setlength{\abovecaptionskip}{-5pt}
	\setlength{\belowcaptionskip}{-10pt}
	\includegraphics[width=1.0\linewidth]{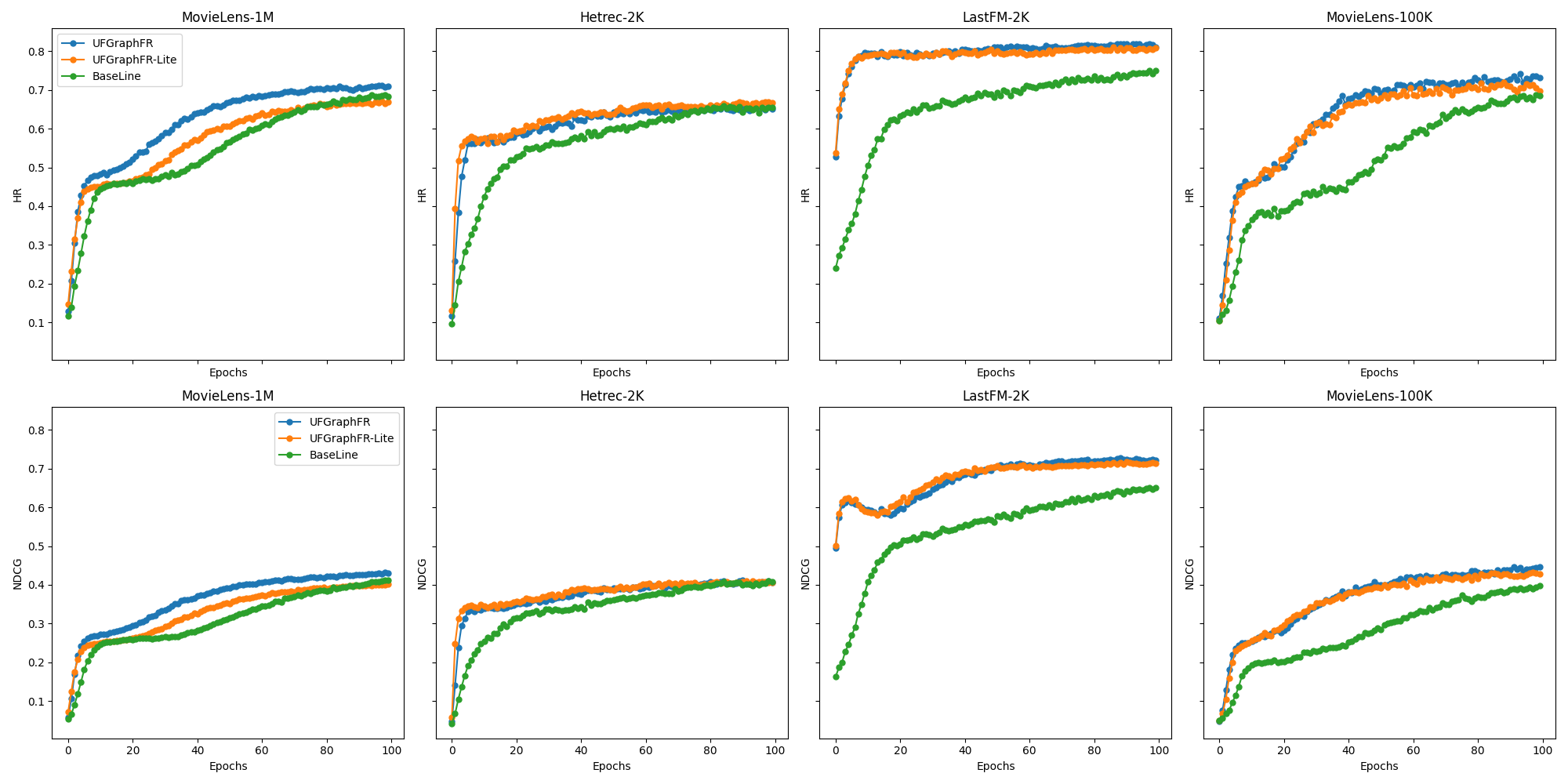}
	\caption{Comparison of model convergence. The horizontal axis is the number of training rounds for federated learning, and the vertical axis is the model performance on both metrics.}
	\label{Convergence}
\end{figure*}

We compared the convergence of our method with the baseline GPFedRec that we mainly compared, and there are two main conclusions: First, as shown in Figure \ref{Convergence}, both UFGraphFR and UFGraphFR-Lite outperformed BaseLine GPFedRec\cite{zhang_gpfedrec_2024} on HR and NDCG on all data sets, suggesting that the two variants were recommended better than the baseline model. Second, UFGraphFR and UFGraphFR-Lite converge relatively quickly on all datasets, especially within the first 50 Epochs.
In addition, we can see that our method converges quickly on all four datasets and the final result is better than the baseline model GPFedRec. There is less interactive data available for each user modeling preference for the Lastfm-2k dataset. Our approach learns personalized item embedding by aggregating highly similar users using their text feature descriptions, alleviating the difficulty of local personalization modeling and accelerating convergence. Secondly, we introduced a Transformer into the model, which can better model the user interaction sequence to speed up convergence.

\section{Conclusion}
In this paper, we propose a novel personalized federated recommendation framework, UFGraphFR, which constructs a user relationship graph based on the user's textual feature descriptions to capture collaborative signals under strict data privacy constraints. To protect user privacy, we design a joint embedding layer that transforms locally embedded user features into low-dimensional representations. Crucially, the server utilizes only the weights of this layer to construct user graphs, without directly accessing or aligning raw user interaction data, upholding a fundamental architectural privacy principle of federated learning. For stronger, quantifiable guarantees, future work will aim to incorporate cryptographic techniques or formal differential privacy analysis into the graph construction and aggregation process.

Furthermore, we integrate a Transformer-based module to capture the temporal dependencies within user-item interaction sequences. Extensive experimental results on four benchmark datasets demonstrate that UFGraphFR significantly outperforms existing state-of-the-art centralized and federated recommendation models, including the strong baseline GPFedRec, in terms of both HR@10 and NDCG@10 metrics. This validates the effectiveness of leveraging semantic relationships derived from user textual features to overcome the user isolation problem inherent in traditional federated recommendation systems.

\subsection{Application Prospects and Real-World Implications}
UFGraphFR presents a practical pathway for deploying compute-intensive, privacy-preserving recommendation tasks in real-world, large-scale federated systems. By offloading the heavy computations—specifically, the large-scale similarity calculations for graph construction and the subsequent graph convolution operations for global aggregation—to a powerful central server or High-Performance Computing (HPC) cluster, the framework aligns with a hybrid computing model. In this model, edge devices (client devices) handle lightweight local training and data collection, ensuring data locality and user privacy, while the centralized supercomputing infrastructure tackles the complex, global integration and relationship mining that are infeasible on resource-constrained clients. The demonstrated performance gains justify the investment in HPC resources for such privacy-aware AI tasks, bridging the gap between advanced federated learning and the scalable, high-throughput requirements of modern supercomputing environments.

\subsection{Limitations and Future Work}
Despite its superior accuracy, UFGraphFR currently incurs higher computational and communication overhead compared to simpler models like GPFedRec, primarily due to the graph construction and aggregation steps. This points to important future directions for improving model efficiency, such as incorporating lightweight Transformer variants, dynamic or periodic graph updating strategies (as preliminarily explored with UFGraphFR-Lite), or more efficient graph neural network architectures.

Regarding privacy, while our framework prevents raw data exposure, the transmission of user embedding weights to the server for graph construction, although derived from local private data, could still pose potential inference risks. To further fortify the system's security, future work will explore integrating advanced cryptographic techniques such as homomorphic encryption for secure similarity computation or applying refined differential privacy mechanisms to the uploaded parameters, thereby enhancing the privacy guarantees during server-side processing.

Finally, the current model focuses solely on user-side textual features for semantic enhancement. A promising direction for future work is to introduce item-side textual features (e.g., product descriptions, article abstracts), enabling bidirectional semantic modeling. This could further improve recommendation quality, cold-start performance, and interpretability, all within the privacy-preserving federated setting.

\bibliographystyle{bst/sn-mathphys-num}

\bibliography{cet3.bib}

\newpage

\section*{Declarations}

\subsection*{Funding}
Not applicable.

\subsection*{Conflict of interest}
The authors declare that they have no known competing financial interests or personal relationships that could have appeared to influence the work reported in this paper.

\subsection*{Ethics approval}
Not applicable.

\subsection*{Availability of data and materials}
All datasets can be obtained from the code repository.
\newpage
\appendix
\section{Appendix: Experimental Settings}
\label{sec:appendix}

This appendix provides the complete experimental parameter configurations, code implementation details, and reproducibility verification methods for the UFGraphFR model.

\subsection{Federated Learning Core Parameters}
\label{subsec:fl_params}

The parameter configurations for federated learning are shown in the table below. These parameters determine the basic framework of federated training, including core settings such as training epochs and client sampling strategies, which directly affect the model's convergence speed and communication overhead.

\begin{table}[h]
\centering
\caption{Federated Learning Core Parameters}
\label{tab:fl_params}
\begin{tabular}{llll}
\hline
\textbf{Parameter} & \textbf{Type} & \textbf{Default} & \textbf{Description} \\
\hline
\texttt{alias} & str & 'UFGraphFR' & Experiment alias \\
\texttt{num\_round} & int & 100 & Total federated learning rounds \\
\texttt{local\_epoch} & int & 1 & Local training epochs per round \\
\texttt{clients\_sample\_ratio} & float & 1.0 & Client sampling ratio \\
\texttt{update\_round} & int & 1 & Model update round \\
\hline
\end{tabular}
\end{table}

\subsection{Graph Construction Parameters}
\label{subsec:graph_params}

Graph construction parameter configurations are shown in the following table.

\begin{table}[h]
\centering
\caption{Graph Construction Parameters}
\label{tab:graph_params}
\begin{tabular}{llll}
\hline
\textbf{Parameter} & \textbf{Type} & \textbf{Default} & \textbf{Description} \\
\hline
\texttt{neighborhood\_size} & int & 0 & Neighborhood size (0 means unlimited) \\
\texttt{neighborhood\_threshold} & float & 1.0 & Neighborhood similarity threshold \\
\texttt{mp\_layers} & int & 1 & Message passing layers \\
\texttt{similarity\_metric} & str & 'cosine' & Similarity metric \\
\texttt{pre\_model} & str & 'USE' & Pre-trained model \\
\texttt{embed\_dim} & int & 100 & Embedding dimension \\
\hline
\end{tabular}
\end{table}

\subsection{Model Architecture Parameters}
\label{subsec:model_params}

Model architecture parameter configurations are shown in the following table.

\begin{table}[h]
\centering
\caption{Model Architecture Parameters}
\label{tab:model_params}
\begin{tabular}{llll}
\hline
\textbf{Parameter} & \textbf{Type} & \textbf{Default} & \textbf{Description} \\
\hline
\texttt{latent\_dim} & int & 32 & Latent vector dimension \\
\texttt{layers} & str & '64, 32, 16, 8' & MLP layers (comma-separated) \\
\texttt{l2\_regularization} & float & 0.0 & L2 regularization coefficient \\
\texttt{dp} & float & 0.2 & Differential privacy noise figure \\
\texttt{reg} & float & 1.0 & Regularization coefficient \\
\texttt{use\_transfermer} & bool & True & Whether to use transfer learning \\
\texttt{use\_jointembedding} & bool & True & Whether to use joint embedding \\
\hline
\end{tabular}
\end{table}

\subsubsection{Model Architecture Details}
\label{subsubsec:model_details}

\begin{itemize}
    \item \textbf{Fractional Function Module}: Three-layer MLP architecture, structure: $64 \rightarrow 32 \rightarrow 8 \rightarrow 1$
    \item \textbf{Text Embedding Processing}: Load USE pre-trained model via MediaPipe, text embedding dimension is 100
    \item \textbf{Joint Embedding Layer}: Output dimension is 32, consistent with \texttt{latent\_dim}
    \item \textbf{User MLP Layer}: Two-layer MLP architecture $32 \rightarrow 64 \rightarrow 32$, consistent with Transformer block's feed-forward network architecture
\end{itemize}

\subsection{Training Optimization Parameters}
\label{subsec:optim_params}

Training optimization parameter configurations are shown in the following table.

\begin{table}[h]
\centering
\caption{Training Optimization Parameters}
\label{tab:optim_params}
\begin{tabular}{llll}
\hline
\textbf{Parameter} & \textbf{Type} & \textbf{Default} & \textbf{Description} \\
\hline
\texttt{lr} & float & 0.1 & Learning rate \\
\texttt{lr\_eta} & int & 80 & Learning rate decay steps \\
\texttt{batch\_size} & int & 256 & Batch size \\
\texttt{optimizer} & str & 'sgd' & Optimizer \\
\texttt{num\_negative} & int & 4 & Number of negative samples \\
\hline
\end{tabular}
\end{table}

\end{document}